\documentclass{article} 
\usepackage{iclr2025_conference,times}

\usepackage{amsmath,amsfonts,bm}

\def\eqref#1{equation~\ref{#1}}

\def\1{\bm{1}}

\DeclareMathAlphabet{\mathsfit}{\encodingdefault}{\sfdefault}{m}{sl}
\SetMathAlphabet{\mathsfit}{bold}{\encodingdefault}{\sfdefault}{bx}{n}

\usepackage{hyperref}
\usepackage{url}
\usepackage{algorithm}
\usepackage{amssymb}
\usepackage{algpseudocode} 
\usepackage{chngcntr}
\counterwithout{algorithm}{section}
\usepackage{booktabs}
\usepackage{multirow}
\usepackage{appendix}
\usepackage{pdflscape}
\usepackage{graphicx}
\usepackage{amsmath,amssymb}
\usepackage{mathtools}
\usepackage{amsthm}
\usepackage{enumitem}
\usepackage{tabularx}
\usepackage{tabularx,array,booktabs}
\newcolumntype{P}[1]{>{\raggedright\arraybackslash}p{#1}}

\title{Deep Global Clustering for Hyperspectral Image Segmentation: Concepts, Applications, and Open Challenges}

\author{
Yu-Tang Chang \quad Pin-Wei Chen \quad Shih-Fang Chen \\
Department of Biomechatronics Engineering \\
National Taiwan University \\
\texttt{\{b05611038,sfchen\}@ntu.edu.tw}
}

\iclrfinalcopy 


\begin{document}

\maketitle
\begin{abstract}
Hyperspectral imaging (HSI) analysis faces computational bottlenecks due to massive data volumes that exceed available memory. 
While foundation models pre-trained on large remote sensing datasets show promise, their learned representations often fail to transfer to domain-specific applications like close-range agricultural monitoring where spectral signatures, spatial scales, and semantic targets differ fundamentally. 
This report presents Deep Global Clustering (DGC), a conceptual framework for memory-efficient HSI segmentation that learns global clustering structure from local patch observations without pre-training. 
DGC operates on small patches with overlapping regions to enforce consistency, enabling training in under 30 minutes on consumer hardware while maintaining constant memory usage. 
On a leaf disease dataset, DGC achieves background-tissue separation (mean IoU 0.925) and demonstrates unsupervised disease detection through navigable semantic granularity. 
However, the framework suffers from optimization instability rooted in multi-objective loss balancing—meaningful representations emerge rapidly but degrade due to cluster over-merging in feature space. 
We position this work as intellectual scaffolding: the design philosophy has merit, but stable implementation requires principled approaches to dynamic loss balancing. 
Code and data are available at \url{https://github.com/b05611038/HSI_global_clustering}.
\end{abstract}

\section{Introduction}
\label{sec:1}
Hyperspectral imaging (HSI) captures hundreds of narrow spectral bands per pixel, adding spectral detail to spatial information and enabling field monitoring and crop assessment in precision agriculture \citep{khan2022101678}. 
Unlike RGB images with three channels, HSI extends to hundreds of wavelength-specific channels (e.g., 400 nm, 402 nm, ..., 1000 nm). 
Critically, HSI channels reflect chemical compositions sensed naturally by the camera \citep{wang2016}, providing physically meaningful features rather than learned abstractions from convolutional layers.
While HSI data may pass through the same computational operations as deep learning features, the practical bottleneck lies elsewhere: data transfer between disk, RAM, and VRAM. 
A single HSI cube (1000 × 1000 pixels, 301 bands) occupies over 1 GB, creating difficulties at every processing stage. 
For datasets containing tens or hundreds of cubes, loading entire datasets into memory becomes impractical.
This technical report extends our conference paper presented at ACPA 2025 \citep{chang2025dgc}, focusing on efficient feature extraction methods for HSI datasets containing millions to billions of pixels.

To prevent misunderstanding the nature of HSI spectral channels, we must clarify an important distinction. 
The wavelength ranges captured by HSI cameras (e.g., 400–1000 nm, or up to 1900 nm in some systems) do not directly reveal the molecular composition of samples—that would require mid-infrared bands \citep{rangle2024}, which are strongly affected by water vapor in practical settings. 
However, this does not diminish the value of visible and near-infrared HSI. 
On the contrary, many objects are readily distinguishable in spectral space even without spatial information. 
For instance, crops in agricultural HSI data can often be separated from backgrounds (tables, conveyors) using spectral signatures alone.
This unique property has profound implications for HSI modeling: treating spectral channels as simply scaled versions of learned features—analogous to treating depth maps as auxiliary features—oversimplifies what HSI represents. 
Because entities can sometimes be detected from spectral information alone, any architecture for HSI analysis must carefully balance spectral and spatial features rather than defaulting to spatial-dominant designs inherited from RGB computer vision.

This study focuses on learning representations that balance spectral and spatial information efficiently. 
In deep learning (DL), "representation" refers to the output at any stage of a neural network, typically the vectors from later layers. These numerical vectors must connect back to real entities for practical use.
When networks receive explicit signals to guide entity recognition, this is supervised learning; when they discover structure without such guidance, this is unsupervised learning—the focus of this work.
How can we realize unsupervised learning for HSI in practice? Foundation models with self-supervised pre-training offer one promising direction, recently exemplified by Hypersigma, HyperSL, and Spectralearth \citep{braham2025, kong2025, wang2025}. 
By scaling transformer pre-training—following successful patterns in other domains—these models outperform methods developed on single datasets. 
However, this approach faces limitations when dealing with unknown sources and entities absent from pre-training data. 
Not every research group can afford the computational cost of large model pre-training or fine-tuning, nor navigate the complexities of hyperparameter tuning at scale. 
Moreover, publicly available HSI datasets are dominated by remote sensing applications, leaving gaps in other domains such as food quality assessment and close-range agricultural monitoring where HSI is also widely used but data remains scarce and domain-specific.

This study deeply examines how differences manifest in HSI data and proposes an efficient algorithm for capturing meaningful representations. 
The core concept—Deep Global Clustering (DGC)—dynamically approximates dataset-level clustering through a representation function that accesses only local patches from individual HSI cubes. 
While DGC exhibits complex learning dynamics that prevent robust convergence, its efficiency enables rapid exploration: successful cases complete training in under 30 minutes on a single GPU with approximately 10 GB VRAM.
We do not claim DGC as a production-ready solution applicable across contexts. 
Rather, we position the conceptual framework—learning global semantic structure from local observations while respecting HSI's spectral-spatial duality—as the durable contribution. 
Both successful and failed cases are reported with diagnostic analysis of instability sources. 
Our hope is that this honest documentation aids researchers developing more robust, generalizable, and efficient methods for HSI representation learning, whether by resolving DGC's optimization challenges or by adopting its design principles in alternative architectures.

\section{Design Philosophy}
\label{sec:2}
Revisiting the meaning of numerical vector representations: how do we distinguish them as different? 
The answer is straightforward—vectors have different values. 
Given any mapping function, the basic unit (pixels in HSI) can be distinguished when spectra differ, ideally while also considering neighboring context. 
This intuition resembles RGB image processing, but what distinguishes HSI at a more abstract level?
In conventional images, a pixel appears different primarily when it contrasts with its neighbors. 
Classical methods like SuperPixels and Histogram of Oriented Gradients (HOG) operationalize this concept by grouping spatially coherent regions or capturing local gradient patterns \citep{petrou2010}.
While DL has largely superseded these hand-crafted features in performance, the underlying principles—spatial coherence and local contrast—remain encoded in learned convolutional filters \citep{zeiler2014}.

\begin{figure}[t]
\centering
\includegraphics[width=0.9\linewidth]{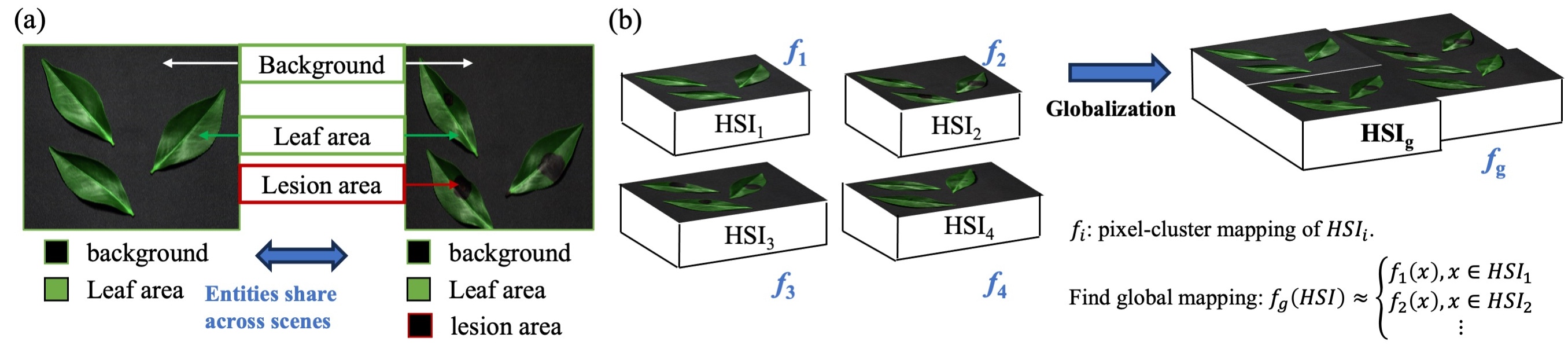}
\caption{Representation of leaf dataset. (a) Global entites shares across two HSI cube. (b) Construction of HSI-DGC.}
\label{fig:1}
\end{figure}

In HSI, spectral features must play a more critical role in representation learning. 
Even when pixels appear similar in pseudo-RGB visualization, their spectral signatures naturally distinguish them if they belong to different entities. 
However, spectral features cannot be used directly—like RGB images, HSI data suffers from illumination variation across acquisitions.
Here lies the critical insight: a group of pixels appears different from neighbors because they are genuinely different in their learned representations. 
Humans perceive a darker region with specific boundaries; in vector space, these manifest as distinct numerical values. 
Extending this concept to the global level (dataset-level): representations of the same entity type should inherit local distinctions while being recognized as a consistent global category (Fig.~\ref{fig:1}a). 

To construct this shared feature space, we need "memory"—anchors that convey differences between grouped pixels. 
Anchors are necessary because human visual cognition does not operate at the pixel level (too noisy and detailed); instead, we perceive clusters of similar objects. 
Clustering thus becomes the natural mechanism for realizing memory in feature space. 
More cluster centers enable finer-grained abstraction of entity types. If every local clustering function can approximate the global one, unsupervised learning from this abstraction produces representations that align with natural human perception.
Formally, this requires finding a global clustering function $f_g$ such that $f_g(HSI) \approx \{f_1(x), x \in HSI_1; f_2(x), x \in HSI_2; ...\}$, where each local function $f_i$ operates on individual HSI cubes (Fig.~\ref{fig:1}b). 
This resembles classical clustering mechanically (assign pixels to nearest centroid), but with an important property: not all K clusters need to be active in every scene. 
A dataset with only healthy leaves may only activate background and tissue clusters, leaving the lesion cluster unused until infected samples appear.
While this may seem impractical mathematically, DGC attempts to realize this concept efficiently through patch-based approximation.

\section{Deep Global Clustering (DGC)}
\label{sec:3}
\subsection{Architecture and Latent Space Refinement}
\label{sec:3.1}

\begin{figure}[t]
\centering
\includegraphics[width=1.0\linewidth]{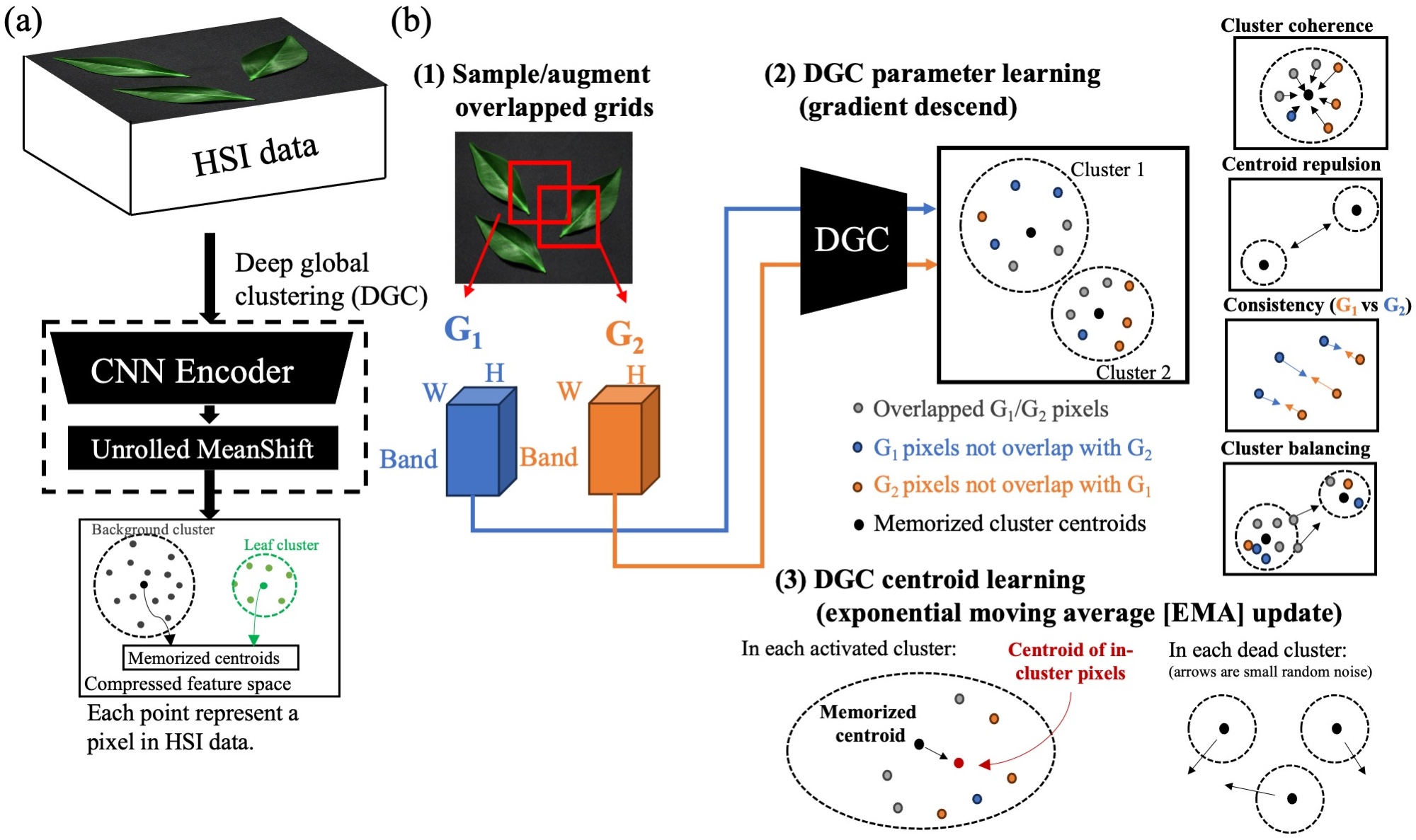}
\caption{Deep Global Clustering (DGC) architecture and training workflow. (a) Architecture: hybrid 1D/2D CNN encoder compresses HSI to 32-dimensional feature space, followed by unrolled mean-shift clustering using memorized centroids. (b) Training procedure: overlapping patches enable learning global cluster structure from local observations through four-term loss optimization and EMA centroid updates.}
\label{fig:2}
\end{figure}

DGC consists of two parametric components: a CNN feature encoder and memorized cluster centroids (Fig.~\ref{fig:2}a). 
The encoder uses a hybrid 1D/2D architecture to balance spectral compression and spatial context. 
Three 1D convolutional layers (kernel size 9) compress the spectral dimension from 301 bands to 32-dimensional embeddings, followed by two 2D convolutional layers (kernel size $3 \times 3$) that incorporate local spatial information. 
All layers use ReLU activation and maintain 32 output channels throughout.
After mapping HSI pixels to this 32-dimensional feature space, an unrolled mean-shift module with five iterations refines the pixel clustering \citep{comaniciu2004}. 
Mean-shift aggregates nearby pixels in feature space while separating distant ones, effectively smoothing the cluster assignments. 
Pixels are then labeled by their nearest memorized centroid (using L2 distance), producing pseudo-segmentation maps.
The learned clusters function as abstract entity types that can be manually aggregated for downstream applications. 
For instance, if DGC learns K=16 clusters and visual inspection reveals clusters 0 and 1 both represent background, they can be merged into a single semantic category. 
While automatic aggregation through neighbor analysis is theoretically possible, this study focuses on manual semantic assignment as a more interpretable and controllable workflow.

\subsection{Learning Strategy}
\label{sec:3.2}

DGC training proceeds in two stages: grid preparation and two-stage optimization (Fig.~\ref{fig:2}b). 

\paragraph{Grid Preparation.}

DGC's core principle is learning global abstractions from local differences.
To operationalize this, two partially overlapping grids ($G_1$, $G_2$) are sampled from the same HSI cube at each training step. 
This overlap creates three pixel subsets with distinct roles:
\begin{itemize}
\item \textbf{Overlapping pixels} ($G_1 \cap G_2$): Must receive consistent cluster assignments despite being processed through different spatial contexts, enforcing global consistency.
\item \textbf{Non-overlapping pixels} ($G_1 \setminus G_2$ and $G_2 \setminus G_1$): Provide independent samples for learning cluster structure within each grid.
\end{itemize}

This differs fundamentally from self-supervised learning approaches that apply different transformations to the same image \citep{gui2024}. 
SSL seeks invariance to augmentations; DGC seeks consistency across spatial regions. 
The overlapping pixels serve as "anchors" connecting local clustering results—if $G_1$ and $G_2$ assign overlapping pixels to consistent clusters despite seeing different spatial neighborhoods, the learned representation captures global structure.
While three or more overlapping grids could strengthen this constraint, two grids provide the minimum configuration needed: one pair of overlapping regions to enforce consistency, plus non-overlapping regions to prevent the model from simply memorizing fixed assignments. 
This balances computational efficiency with the conceptual requirement of linking local observations to global clustering.

\paragraph{DGC Learning: Gradient Descent for Feature Encoding and Exponential Moving Average for Centroids}

DGC learns the CNN feature encoder and cluster centroids through two alternating phases: (I) gradient descent updates the encoder to construct a better embedding space given fixed centroids, and (II) exponential moving average (EMA) repositions centroids based on the current embedding space. These phases are mutually dependent: phase (I) produces better clustering geometry given current centroids, while phase (II) moves centroids to positions that better represent pixel groupings in the learned space.

This learning strategy optimistically assumes the feature space can be constructed into a good geometry from random initialization. In practice, this does not occur without intervention. Therefore, phase (I) includes an unconventional regularization term in the loss function to handle poor initialization. We acknowledge this term affects convergence and likely contributes to instability in some scenarios. However, without this adjustment, DGC fails to learn any semantic structure regardless of training duration. We view this as an inevitable tradeoff between initialization robustness and optimization stability.

In phase (I), DGC optimizes four complementary objectives on overlapping patch pairs $(G_1, G_2)$:

\textbf{(1) Compactness}: Pulls pixels toward their assigned cluster centers in embedding space:
$$\mathcal{L}_{\text{comp}} = \frac{1}{BN} \sum_{b,i,k} p_{bi}^{(k)} (1 - \cos(z_{bi}, c_k))$$
where $p_{bi}^{(k)}$ is the soft assignment probability and $\cos(\cdot, \cdot)$ denotes cosine similarity between normalized embeddings.

\textbf{(2) Orthogonality}: Encourages cluster centroids to be diverse by penalizing their pairwise correlations:
$$\mathcal{L}_{\text{orth}} = \sum_{i \neq j} (c_i^\top c_j)^2 = \|CC^\top - I\|_F^2$$

\textbf{(3) Balance + Uniform Assignment}: Two complementary terms prevent cluster collapse. The balance term maximizes entropy of marginal cluster usage:
$$\mathcal{L}_{\text{bal}} = -\sum_{k=1}^{K} \bar{p}_k \log \bar{p}_k, \quad \bar{p}_k = \frac{1}{BN}\sum_{b,i} p_{bi}^{(k)}$$

The uniform assignment term addresses cluster imbalance by encouraging a uniform distribution of pixels across all $K$ clusters. It constructs pseudo-labels targeting exactly $\lfloor M/K \rfloor$ pixels per cluster and minimizes cross-entropy against the model's soft predictions:
$$\mathcal{L}_{\text{unif}} = -\frac{1}{M} \sum_{i=1}^{M} \log p_i^{(y_i^*)}$$
where $y_i^* \in \{1,...,K\}$ are pseudo-labels designed to enforce balanced cluster usage. This provides strong redistribution pressure to escape initialization collapse where all pixels are assigned to a single cluster. The practical implementation of pseudo-label construction involves sampling strategies that approximate uniform target distributions while remaining computationally efficient.

\textbf{(4) Consistency}: Enforces similar cluster distributions across overlapping patches via symmetric KL divergence:
$$\mathcal{L}_{\text{cons}} = \frac{1}{2M}\sum_i [\text{KL}(p_i^{(1)} \| p_i^{(2)}) + \text{KL}(p_i^{(2)} \| p_i^{(1)})]$$

The total loss combines these terms: 
$$\mathcal{L} = \mathcal{L}_{\text{comp}}^{(1)} + \mathcal{L}_{\text{comp}}^{(2)} + \lambda_{\text{unif}} \mathcal{L}_{\text{unif}} + \lambda_{\text{orth}} \mathcal{L}_{\text{orth}} + \lambda_{\text{bal}} \mathcal{L}_{\text{bal}} + \lambda_{\text{cons}} \mathcal{L}_{\text{cons}}$$
where superscripts denote the two crops. 

This integration enforces pixel-level features to group with local neighbors while maintaining repulsion between memorized centroids, mimicking the cognitive process of glimpsing complex patterns for simple visual grouping. By operating directly on pixel-level features with limited spatial context, DGC constructs a feature space fundamentally different from foundation models. We note that transformer-based architectures are poorly suited for this conceptual framework due to extensive pixel interactions in self-attention mechanisms. The structural inductive bias of CNNs—local receptive fields and limited spatial mixing—better aligns with the principle of learning global structure from local observations.

\textbf{Centroid Updates}: After each gradient step, cluster centers are updated via exponential moving average (EMA) based on the weighted mean of assigned pixel embeddings:
$$c_k \leftarrow \alpha c_k + (1-\alpha) \frac{\sum_i p_i^{(k)} z_i}{\sum_i p_i^{(k)}}$$
Dead clusters (those with assignment mass below threshold) receive small random perturbations to encourage reactivation. All centers are then L2-normalized to maintain unit norm.

This second stage serves as position refinement for cluster centroids. While stage I (gradient descent) optimizes the feature space geometry, it provides no guarantee that the memorized centroids occupy their true geometric centers in that space. Stage II prevents centroids from drifting away from appropriate regions—ensuring they remain close to their assigned pixel clusters in feature space, even if not at the exact mathematical centroid. This refinement is essential to the DGC framework: centroids must accurately represent their clusters to enable consistent global clustering across local patch observations.

\subsection{Implementation Details}
\label{sec:3.3}

For specific hyperparameter configurations, please refer to the GitHub repository where default settings correspond to those used in this report.

\textbf{Efficiency through patch-based learning.} DGC achieves memory efficiency by operating on small patches rather than full HSI cubes. Each training step samples two $64\times64$ patches from a single $1000\times1000$ cube ($\approx250\times$ reduction in pixels per iteration), with partial overlap to enforce global consistency via $\mathcal{L}_{\text{cons}}$. Once a cube is loaded, multiple patch pairs are sampled before advancing to the next cube, increasing effective iteration counts by $10\times$ to $100\times$ without additional disk I/O. This reuse strategy addresses the primary I/O bottleneck: loading a 1GB cube from disk requires several seconds even with SSDs, but each loaded cube yields dozens of training iterations.

Furthermore, asynchronous I/O with dual memory buffers—one for training while another handles loading, then switching when I/O completes—can push iteration throughput to GPU computational limits. We do not provide comprehensive comparison between synchronous and asynchronous DGC variants, as the speedup is self-evident. However, our experiments with asynchronous DGC reveal significant stability issues affecting convergence, which we discuss in \S\ref{sec:5}.

On a consumer GPU (NVIDIA RTX 4080) with batch size 4, successful synchronous training completes in under 30 minutes for the 86-cube leaf dataset, using approximately 10GB VRAM.
The asynchronous implementation is included in the GitHub repository as a reference for readers attempting to improve stability or seeking deeper understanding of the conceptual framework.

\section{Application: Leaf Disease HSI Dataset}
\label{sec:4}

\subsection{Dataset Description}
\label{sec:4.1}

HSI data were collected using a Hyperspec MV.X hyperspectral camera (Headwall Photonics Inc.; Bolton, MA, USA). 
Each cube contains $1000\times1000$ pixels across 301 spectral bands spanning 400–1000 nm with 2 nm resolution. 
Scenes were captured on a black conveyor belt and contain three entity types: background (conveyor surface), healthy leaf tissue, and diseased tissue affected by brown blight. 
The dataset comprises 86 samples total: 22 healthy leaves and 64 infected leaves exhibiting visible lesions.

\begin{figure}[t]
\centering
\includegraphics[width=1.0\linewidth]{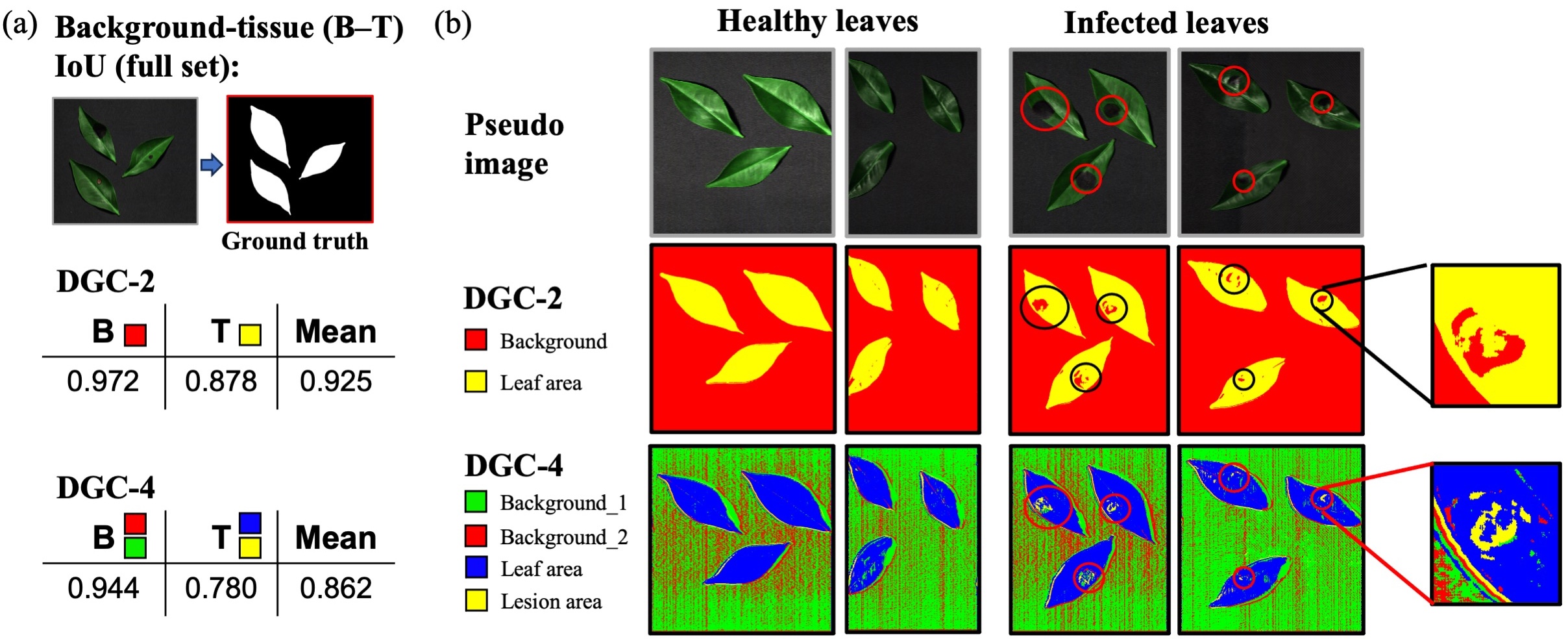}
\caption{DGC clustering results on leaf dataset. (a) Background-tissue (B–T) IoU scores for DGC-2 and DGC-4 evaluated against manual annotations. (b) Representative pseudo-segmentation outputs on healthy and infected leaves.}
\label{fig:3}
\end{figure}

To evaluate DGC's ability to extract abstract entity-level features, we manually annotated pixel-level masks for leaf area and background regions, providing ground truth for background-tissue (B–T) binary separation (Fig.~\ref{fig:3}a). 
The primary evaluation metric is Intersection over Union (IoU), computed separately for background (B) and tissue (T) classes, with mean IoU reported as the overall performance measure. 
This metric assesses whether DGC's unsupervised clustering captures the fundamental spectral distinction between background and biological tissue without supervision.

\subsection{Synchronous DGC Results}
\label{sec:4.2}

We report results for two configurations: DGC-2 (two clusters) and DGC-4 (four clusters). 
Regardless of whether DGC learns semantically meaningful abstractions, human judgment remains necessary to interpret the discovered clusters (Fig.~\ref{fig:3}b). 
Through visual inspection of predicted classes, we assigned semantic labels: for DGC-2, one cluster corresponds to background and the other to leaf tissue. 
DGC-4 presents greater complexity—one learned cluster appears sparsely across the dataset and, upon inspection, represents diseased tissue regions.

DGC-2 produces clean background-tissue separation, though some lesion pixels near leaf boundaries and shadows are misclassified as background. 
In the pseudo-RGB visualization, these lesions appear nearly black with spectra similar to the conveyor background, making spectral-based confusion understandable for an unsupervised model. 
DGC-4 provides finer granularity: background separates into flat surface versus textured regions, while tissue splits into healthy and lesion-affected areas. 
Using manual annotations to evaluate B–T separation, DGC-2 achieves IoU 0.972 (background), 0.878 (tissue), and 0.925 mean; DGC-4 achieves IoU 0.944 (background), 0.780 (tissue), and 0.862 mean (Fig.~\ref{fig:3}a).

These results demonstrate the navigable granularity principle: DGC-2 provides coarse entity-level separation with higher accuracy, while DGC-4 reveals finer semantic structure at the cost of lower B–T scores. 
Notably, disease areas form coherent clusters without any supervision or prior knowledge of lesion locations, demonstrating unsupervised disease detection capability. 
The degradation in B–T IoU for DGC-4 reflects the model's allocation of representational capacity toward finer distinctions rather than failure—the semantic information is richer, though the binary B–T metric does not capture this added value.

\subsection{Asynchronous DGC Results}
\label{sec:4.3}

\begin{figure}[t]
\centering
\includegraphics[width=0.9\linewidth]{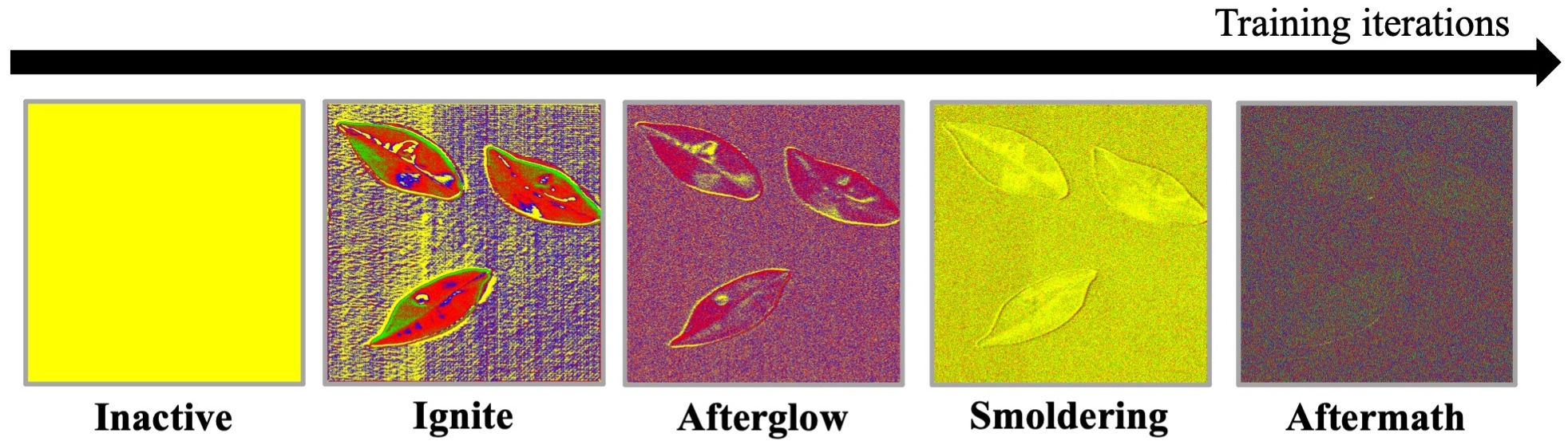}
\caption{Training evolution of Asynchronous DGC-4. Sequential pseudo-segmentation outputs show five phases: inactive, ignite, afterglow, smoldering, aftermath.}
\label{fig:4}
\end{figure}

Asynchronous DGC exhibits severe hyperparameter sensitivity, making stable training difficult to achieve. 
We therefore do not report quantitative B–T separation performance for this variant. 
However, the training dynamics reveal important insights into DGC's learning process. 
Under certain hyperparameter configurations, semantic patterns do emerge, but they fade rapidly—typically persisting for fewer than 1000 iterations before collapsing (Fig.~\ref{fig:4}; Async-DGC-4). 
The appearance of meaningful clustering resembles fireworks: a long ignition period followed by a brief moment of clarity before fading. 
We label five observable stages: \textbf{Inactive} (all pixels collapsed to one cluster), \textbf{Ignite} (semantic patterns suddenly emerge—the desired state), \textbf{Afterglow} (cluster mixing begins in feature space), \textbf{Smoldering} (over-merging becomes severe), and \textbf{Aftermath} (complete collapse to noise). 
The ignite phase represents the brief moment when DGC achieves what we want, but it degrades too quickly to be useful.

This visualization is not self-deprecation but rather diagnostic evidence of DGC's learning phases: from biased initialization (pixels near one centroid) to over-merging of abstract clusters in feature space. Without the uniform assignment term, DGC remains stuck in the inactive phase indefinitely—patterns never emerge. 
However, this same mechanism that breaks cluster collapse eventually destroys cluster boundaries entirely, leading to the final noise-like pseudo-segmentation.
The evolutionary pattern suggests DGC's failure lies not in the conceptual framework itself, but in determining the proper optimization endpoint. 
The model briefly achieves meaningful representations but lacks a mechanism to recognize and preserve this state. Further discussion of this optimization challenge appears in Section~\ref{sec:5}.

\section{Discussion}
\label{sec:5}

DGC is not a successful algorithm in the conventional sense, but rather a well-explored conceptual framework. 
We identify three potential failure modes: loss function design, grid sampling strategy, and termination mechanism.

\paragraph{Loss Function Design.} 
Conventionally, loss functions in deep learning serve as surrogates for measuring target achievement. 
Readers might attribute DGC's failure to excessive complexity (too many terms), but we believe the core problem is not multi-objective learning per se. 
Rather, it stems from lacking a principled mechanism to balance contradictory objectives—particularly intra-cluster coherence versus uniform assignment. 
DGC's learning dynamics resemble an evolutionary system seeking equilibrium, analogous to atomic structure: protons and neutrons bind through strong interaction while electrons orbit due to electromagnetic repulsion. 
A stable equilibrium form may exist, but we have not found it in our experiments. 
Exhaustive hyperparameter tuning is clearly inadequate; what is needed is a more sophisticated yet natural integration scheme that balances all terms dynamically rather than through fixed scalar weights.

\paragraph{Grid Sampling Strategy.} 
The synchronous version tolerates a wider hyperparameter range than the asynchronous variant. 
We speculate this occurs because limited data access implicitly creates a beneficial sampling strategy that maintains stable dynamics in the evolutionary system. 
In async-DGC, the model repeatedly accesses the same cube (despite random patch locations) while the memory buffer loads the next batch. 
This repetition may introduce data-level bias that skews cluster representations, breaking equilibrium when switching to newly loaded data.
Given DGC's efficiency objectives, we did not extensively explore spatial sampling uniformity across the 2D image plane. 
However, we believe a low-cost algorithm ensuring quasi-random spatial sampling could stabilize DGC learning without sacrificing computational efficiency.

\paragraph{Termination Mechanism.} 
Human-in-the-loop evaluation is essential to DGC because semantic standards vary across applications. 
We originally envisioned DGC's evolution producing a sequence of valid clusterings from which users could select. 
The merging strategy for learned clusters remains an important direction for future exploration.
Regarding the optimization endpoint, we believe it is inappropriate to define a universal stopping criterion based on internal metrics.
We tested various measures—entropy and variance of segmentations, mutual information between consecutive iterations—but none reliably indicated when to terminate. 
A more promising direction may be ensuring DGC traverses a better optimization trajectory where all intermediate states represent valid clustering solutions, rather than seeking a single "correct" endpoint.

\paragraph{Sparse Cluster Activation.}
A final reminder: DGC does not aim to find K-means-like global clustering in the traditional sense. 
The core principle is that spectrally similar pixels should group together upon first inspection (mapping to feature space). 
A commonly overlooked aspect of this process is the \textbf{existence of a sparse selection mechanism to activate only relevant clusters (concepts)}. 
Although not explicitly defined in the computation, we expect DGC to learn feature groups bottom-up, entirely from data without priors—analogous to how humans form concepts when encountering novel objects. 
Not all K clusters need activate simultaneously; rather, only those representing entities present in a given scene should engage.

\section{Conclusion}
\label{sec:6}

This report presents Deep Global Clustering (DGC), a conceptual framework for memory-efficient HSI analysis that learns global semantic structure from local patch observations. 
While the synchronous implementation demonstrates rapid semantic emergence—achieving background-tissue separation (mean IoU 0.925) and unsupervised disease detection within 30 minutes on consumer hardware—the framework suffers from optimization instability rooted in multi-objective loss balancing. 
The asynchronous variant's "firework" behavior reveals that meaningful representations emerge briefly during training but degrade due to over-merging in feature space, lacking a mechanism to recognize and preserve stable states. 
We position this work as intellectual scaffolding: the design philosophy of navigable semantic granularity and sparse cluster activation has merit, but realizing stable implementations requires advances in principled loss balancing and trajectory optimization. 
By honestly documenting both successful cases and failure modes, we hope to accelerate development of robust methods for unsupervised HSI analysis in domain-specific applications where foundation models remain inaccessible.

\section*{Acknowledgments}

We thank Xiu-Rui Lin (Agricultural Chemicals Research Institute, Ministry of Agriculture, Taichung City, Taiwan) for her assistance with data collection and for providing essential background knowledge on leaf diseases and the visual characteristics of infected tissues. 

\section*{Author Contributions}

Yu-Tang Chang: Conceptualization, methodology, software implementation, analysis, and writing. Pin-Wei Chen: Data collection, annotation, and validation. Shih-Fang Chen: Supervision and project administration.

\nocite{*}
\bibliographystyle{iclr2025_conference}
\bibliography{iclr2025_conference}

\begin{thebibliography}{12}
\providecommand{\natexlab}[1]{#1}
\providecommand{\url}[1]{\texttt{#1}}
\expandafter\ifx\csname urlstyle\endcsname\relax
  \providecommand{\doi}[1]{doi: #1}\else
  \providecommand{\doi}{doi: \begingroup \urlstyle{rm}\Url}\fi

\bibitem[Braham et~al.(2025)Braham, Albrecht, Mairal, Chanussot, Wang, and Zhu]{braham2025}
Nassim Ait~Ali Braham, Conrad~M. Albrecht, Julien Mairal, Jocelyn Chanussot, Yi~Wang, and Xiao~Xiang Zhu.
\newblock Spectralearth: Training hyperspectral foundation models at scale.
\newblock \emph{IEEE Journal of Selected Topics in Applied Earth Observations and Remote Sensing}, 18:\penalty0 16780--16797, 2025.
\newblock \doi{10.1109/JSTARS.2025.3581451}.

\bibitem[Chang et~al.(2025)Chang, Chen, and Chen]{chang2025dgc}
Yu-Tang Chang, Pin-Wei Chen, and Shih-Fang Chen.
\newblock Unsupervised hyperspectral image segmentation using deep global clustering.
\newblock In \emph{Proceedings of the 11th Asian-Australasian Conference on Precision Agriculture (ACPA)}, Chiayi, Taiwan, October 2025.
\newblock Extended abstract.

\bibitem[Comaniciu \& Meer(2002)Comaniciu and Meer]{comaniciu2004}
D.~Comaniciu and P.~Meer.
\newblock Mean shift: a robust approach toward feature space analysis.
\newblock \emph{IEEE Transactions on Pattern Analysis and Machine Intelligence}, 24\penalty0 (5):\penalty0 603--619, 2002.
\newblock \doi{10.1109/34.1000236}.

\bibitem[{Flores Rangel} et~al.(2024){Flores Rangel}, {Diaz de León Martínez}, Walter, and Mizaikoff]{rangle2024}
Gabriela {Flores Rangel}, Lorena {Diaz de León Martínez}, Lisa~Sophie Walter, and Boris Mizaikoff.
\newblock Recent advances and trends in mid-infrared chem/bio sensors.
\newblock \emph{TrAC Trends in Analytical Chemistry}, 180:\penalty0 117916, 2024.
\newblock ISSN 0165-9936.
\newblock \doi{https://doi.org/10.1016/j.trac.2024.117916}.
\newblock URL \url{https://www.sciencedirect.com/science/article/pii/S0165993624003996}.

\bibitem[Gui et~al.(2024)Gui, Chen, Zhang, Cao, Sun, Luo, and Tao]{gui2024}
Jie Gui, Tuo Chen, Jing Zhang, Qiong Cao, Zhenan Sun, Hao Luo, and Dacheng Tao.
\newblock A survey on self-supervised learning: Algorithms, applications, and future trends.
\newblock \emph{IEEE Transactions on Pattern Analysis and Machine Intelligence}, 46\penalty0 (12):\penalty0 9052--9071, 2024.
\newblock \doi{10.1109/TPAMI.2024.3415112}.

\bibitem[Khan et~al.(2022)Khan, Vibhute, Mali, and Patil]{khan2022101678}
Atiya Khan, Amol~D. Vibhute, Shankar Mali, and C.H. Patil.
\newblock A systematic review on hyperspectral imaging technology with a machine and deep learning methodology for agricultural applications.
\newblock \emph{Ecological Informatics}, 69:\penalty0 101678, 2022.
\newblock ISSN 1574-9541.
\newblock \doi{https://doi.org/10.1016/j.ecoinf.2022.101678}.
\newblock URL \url{https://www.sciencedirect.com/science/article/pii/S1574954122001285}.

\bibitem[Kong et~al.(2025)Kong, Liu, Bi, Yu, Li, and Chen]{kong2025}
Weili Kong, Baisen Liu, Xiaojun Bi, Changdong Yu, Xinyao Li, and Yushi Chen.
\newblock Hypersl: A spectral foundation model for hyperspectral image interpretation.
\newblock \emph{IEEE Transactions on Geoscience and Remote Sensing}, 63:\penalty0 1--19, 2025.
\newblock \doi{10.1109/TGRS.2025.3566205}.

\bibitem[Petrou \& Petrou(2010)Petrou and Petrou]{petrou2010}
Maria Petrou and Costas Petrou.
\newblock \emph{Image Processing: The Fundamentals}.
\newblock John Wiley \& Sons, Chichester, West Sussex, UK, 2nd edition, 2010.
\newblock ISBN 978-0-470-74586-1.
\newblock \doi{10.1002/9781119994398}.

\bibitem[Wang et~al.(2025)Wang, Hu, Jin, Miao, Yang, Xu, Qin, Ma, Sun, Li, Fu, Chen, Han, Yokoya, Zhang, Xu, Liu, Zhang, Wu, Du, Tao, and Zhang]{wang2025}
Di~Wang, Meiqi Hu, Yao Jin, Yuchun Miao, Jiaqi Yang, Yichu Xu, Xiaolei Qin, Jiaqi Ma, Lingyu Sun, Chenxing Li, Chuan Fu, Hongruixuan Chen, Chengxi Han, Naoto Yokoya, Jing Zhang, Minqiang Xu, Lin Liu, Lefei Zhang, Chen Wu, Bo~Du, Dacheng Tao, and Liangpei Zhang.
\newblock Hypersigma: Hyperspectral intelligence comprehension foundation model.
\newblock \emph{IEEE Transactions on Pattern Analysis and Machine Intelligence}, 47\penalty0 (8):\penalty0 6427--6444, 2025.
\newblock \doi{10.1109/TPAMI.2025.3557581}.

\bibitem[Wang \& Zhao(2016)Wang and Zhao]{wang2016}
Liguo Wang and Chunhui Zhao.
\newblock \emph{Hyperspectral Image Processing}.
\newblock Springer, Berlin, Heidelberg, 2016.
\newblock ISBN 978-3-662-47456-3.
\newblock \doi{10.1007/978-3-662-47456-3}.
\newblock URL \url{https://doi.org/10.1007/978-3-662-47456-3}.

\bibitem[Xie et~al.(2020)Xie, Zhang, Liao, Xia, and Shen]{xie2020}
Yutong Xie, Jianpeng Zhang, Zehui Liao, Yong Xia, and Chunhua Shen.
\newblock Pgl: Prior-guided local self-supervised learning for 3d medical image segmentation.
\newblock \emph{arXiv preprint arXiv:2011.12640}, November 2020.
\newblock URL \url{https://arxiv.org/abs/2011.12640}.

\bibitem[Zeiler \& Fergus(2014)Zeiler and Fergus]{zeiler2014}
Matthew~D. Zeiler and Rob Fergus.
\newblock Visualizing and understanding convolutional networks.
\newblock In \emph{Computer Vision -- ECCV 2014}, pp.\  818--833. Springer, 2014.
\newblock \doi{10.1007/978-3-319-10590-1_53}.

\end{thebibliography}

\end{document}